\documentclass[nohyperref]{article}

\usepackage{microtype}
\usepackage{graphicx}
\usepackage{subfigure}
\usepackage{booktabs} 

\usepackage{hyperref}

\usepackage[accepted]{icml2022}

\usepackage{amsmath}
\usepackage{amssymb}
\usepackage{mathtools}
\usepackage{amsthm}

\usepackage[capitalize,noabbrev]{cleveref}

\theoremstyle{plain}

\theoremstyle{definition}

\theoremstyle{remark}

\usepackage[textsize=tiny]{todonotes}

\icmltitlerunning{Fair and Green Hyperparameter Optimization via Multi-objective and Multiple Information Source Bayesian Optimization}

\begin{document}

\twocolumn[
\icmltitle{Fair and Green Hyperparameter Optimization via Multi-objective and\\ Multiple Information Source Bayesian Optimization}



\icmlsetsymbol{equal}{*}

\begin{icmlauthorlist}
\icmlauthor{Antonio Candelieri}{dems}
\icmlauthor{Andrea Ponti}{dems,oaks}
\icmlauthor{Francesco Archetti}{disco}
\end{icmlauthorlist}

\icmlaffiliation{dems}{Department of Economics, Management and Statistics, University of Milano-Bicocca, Milan, Italy}
\icmlaffiliation{disco}{Department of Computer Science, University of Milano-Bicocca, Milan, Italy}
\icmlaffiliation{oaks}{OAKS srl, Milan, Italy}

\icmlcorrespondingauthor{Antonio Candelieri}{antonio.candelieri@unimib.it}

\icmlkeywords{Fair Machine Learning, Green Machine Learning, Multi-objective Optimization, Multiple Information Source Optimization, Bayesian Optimization}

\vskip 0.3in
]



\printAffiliationsAndNotice{\icmlEqualContribution} 

\begin{abstract}
There is a consensus that focusing only on accuracy in searching for optimal machine learning models amplifies biases contained in the data, leading to unfair predictions and decision supports. Recently, multi-objective hyperparameter optimization has been proposed to search for machine learning models which offer equally Pareto-efficient trade-offs between accuracy and fairness. Although these approaches proved to be more versatile than fairness-aware machine learning algorithms – which optimize accuracy constrained to some threshold on fairness – they could drastically increase the energy consumption in the case of large datasets. In this paper we propose FanG-HPO, a Fair and Green Hyperparameter Optimization (HPO) approach based on both multi-objective and multiple information source Bayesian optimization. FanG-HPO uses subsets of the large dataset (aka information sources) to obtain cheap approximations of both accuracy and fairness, and multi-objective Bayesian Optimization to efficiently identify Pareto-efficient machine learning models. Experiments consider two benchmark (fairness) datasets and two machine learning algorithms (XGBoost and Multi-Layer Perceptron), and provide an assessment of FanG-HPO against both fairness-aware machine learning algorithms and hyperparameter optimization via a multi-objective single-source optimization algorithm in BoTorch, a state-of-the-art platform for Bayesian Optimization.

\begin{small}
\end{small}
\end{abstract}

\section{Introduction}
\label{sec:intro}

\subsection{Rationale and motivations}
A low misclassification/prediction error is not the only performance metric of interest in searching for the most suitable Machine Learning (ML) model to use in a successful decision support application. Additional metrics like \textit{fairness}, \textit{interpretability}, and \textit{privacy} have been increasingly becoming important during last years \citep{barocas2017fairness}. This paper focuses on \textit{fairness}, a desired property of the decision support provided by a ML model: it must not be “biased” towards a specific person or groups of individuals. The topic is widely known as \textit{FairML} \citep{mehrabi2021survey}, with approaches which can be organized into three different families:
(i) \textit{post-processing} to modify a pre-trained model to increase the fairness of its outcomes, (ii) \textit{in-processing} to enforce fairness constraints during training, and (iii) \textit{pre-processing} to modify the data representation and then apply standard ML algorithms \citep{friedler2019comparative}.

These approaches regard the design of \textit{fairness-aware} (or \textit{fair-by-design}) ML algorithms, but they suffer from one or more of the following drawbacks \citep{perrone2021fair}: the intervention performed to deal with biases is (\textit{i}) specific to the model class (e.g., linear models only), (\textit{ii}) limited to a specific definition(s) of fairness, (\textit{iii}) limited to a single, binary sensitive feature, (\textit{iv}) requires access to sensitive feature information at prediction time, and (\textit{v})  results in a randomized classifier that may generate different prediction for the same input at different times.

These considerations have recently led to a new proposition: instead of designing fairness-aware ML algorithm, FairML can be addressed as an hyperparameter optimization (HPO) task which also considers fairness as a metric. Indeed, HPO of a ML algorithm, only driven by misclassification/prediction error minimization, can amplify biases contained in the dataset, leading to an unfair decision support \citep{barocas2017fairness,buolamwini2018gender}. Recently, two types of approaches have been proposed to perform HPO for FairML: (\textit{i})
\textbf{constrained optimization} -- aimed at maximizing accuracy while satisfying some given fairness constraint \citep{perrone2020bayesian,perrone2021fair} -- and (\textit{ii})
\textbf{Multi-objective optimization} -- aimed at maximizing, simultaneously, accuracy and some fairness metric \citep{schmucker2020multi}.

It is important to remark that the first approach could not be applicable in many real-life settings, because suitable fairness constraints could be difficult to be established a-priori. According to this consideration, we have decided to focus our research on the multi-objective optimization approach.

In addition to fairness, this paper also addresses the issue of \textit{energy-efficiency} of an HPO task. Nowadays, it has become crucial to consider the dual role of Artificial Intelligence (AI) and ML in the climate crisis. On the one hand, they can support more sustainable and low-emission decisions, from design to management of critical systems such as smart energy-grids, transportation, healthcare and water utilities, and they can also provide accurate climate change predictions. On the other hand, AI and ML are themselves \textit{energivorous} and, consequently, significant emitters of CO$_2$, leading to the concept of Red-AI \citep{dhar2020carbon}. Prevalence of Red-AI is also quantified in \citep{schwartz2020green}, reporting that the total cost of producing accurate ML models increases linearly with \textit{(i)} the cost of executing the model on a single example, \textit{(ii)} the size of the training dataset and \textit{(iii)} the number of HPO experiments, which controls how many times the model is trained on the dataset. Astonishing results are also reported in \citep{strubell2019energy,hao2019training}, which analysed the training process of many Natural Language Processing (NLP) models to estimate the energy cost in kilowatts required. When these figures are converted into approximate carbon emissions it comes out that the carbon footprint of training a single large NLP model is equal to the amount of CO$_2$ emitted by 125 round-trip flights between New York and Beijing or, equivalently, five American average cars in their lifetimes, including their manufacturing processes. Consequently, the research community has been focusing on the Green-AI topic and also starting to propose novel optimization techniques to make the HPO task “greener” \citep{tornede2021towards}, for instance by using smaller portions of the available databases/datasets, as proposed in the seminal work of \citep{swersky2013multi} to the most recent ones, such as in \citep{klein2017fast,candelieri2021green}.

Bayesian Optimization (BO) is a sample-efficient, sequential, model-based, global optimization method, well-suited for optimizing black-box, expensive, and multi-extremal objective functions \citep{frazier2018bayesian,archetti2019bayesian}. Thanks to its sample-efficiency, BO is the core component of most of the current Automated Machine Learning (AutoML) \citep{hutter2019automated,he2021automl} solutions, both open-source and commercial. BO has been recently extended to also deal with multiple objectives \citep{hernandez2016predictive,paria2020flexible}, as well as multiple information sources with different computational cost can be accessed \citep{ghoreishi2019multi,belakaria2020multi,candelieri2021green,candelieri2021sparsifying,khatamsaz2020efficient}. A special case is when sources can be organized hierarchically depending on their quality of approximation (aka \textit{fidelity}), leading to the so-called multi-fidelity optimization, originally proposed in \citep{kennedy2000predicting}.

\subsection{Contributions}
The main contributions of this paper are:
\begin{enumerate}
    \item A new \textbf{\underline{F}}air \textbf{\underline{an}}d \textbf{\underline{G}}reen Hyperparameter Optimization algorithm, namely \textbf{FanG-HPO}, based on both multi-objective and multiple information source Bayesian Optimization, which offers more flexibility and better computational performance on HPO problems.
    \item A computational assessment of FanG-HPO against both fairness-aware ML algorithms \citep{scutari2021achieving} and fair-HPO performed by using BoTorch \citep{balandat2020botorch}, a state-of-the-art platform for BO.
\end{enumerate}

The rest of the paper is organized as follows: Sect. 2 provides the main background on multi-objective and multiple information source optimization, along with the definition of the fairness metric adopted in the study. In Sect. 3 the FanG-HPO approach is detailed. Sect. 4 describes the experimental setting and reports results. Finally, Sect. 5 provides conclusions and perspectives.

\subsection{related works}

A recent BO related study targeting fairness in ML is \citep{schmucker2020multi} whose proposed approach combines multi-objective and multi-fidelity by building upon Hyperband \citep{li2017hyperband}. Unfortunately the code is not currently available: an implementation was uder review to be included into AutoGluon\footnote{https://auto.gluon.ai/stable/index.html} suite, but it has not been included yet.
Other relevant works are \citep{perrone2021fair,perrone2020bayesian}, where the fairness requirement are modelled as a constraint -- instead of an objective -- in single-objective (i.e., accuracy maximization) HPO.

With respect to fairness-aware ML algorithms, relevant works are \citep{komiyama2018nonconvex,zafar2019fairness,scutari2021achieving}.

Finally, although BO has been extended to deal with multi-objectives \citep{svenson2016multiobjective,feliot2017bayesian,yang2019multi,iqbal2020flexibo,daulton2020differentiable} as well as multiple fidelities and multiple information sources \citep{lam2015multifidelity,poloczek2017multi,ghoreishi2019multi,candelieri2021sparsifying,candelieri2021miso,ariafar2021faster}, there is a significant lack of solutions jointly addressing the two tasks. On the other hand, the research interest on this specific challenge is quickly increasing, especially because its applicability to many other real-life problems than fair and green ML, as demonstrated by very recent works such as \citep{sun2022correlated} and \citep{irshad2021expected}.

\section{Background}
This paper addresses FairML as a multi-objective problem and, at the same time, uses multiple information sources (i.e., a small portion of the large target dataset) to improve energy-efficiency and, consequently, reduce CO$_2$ emissions. Here, we briefly summarize the basic background about multi-objective optimization, fairness metrics and multiple information source optimization.

\subsection{Multi-objective optimization}
Multi-objective optimization (MO) concerns solving problems with more than one objective function to be optimized simultaneously, that is:
\begin{equation}
    \label{eq:mo}
    \underset{\mathbf{x} \in \Omega}{\min}\; \mathbf{f}(\mathbf{x})
\end{equation}

where $\Omega$ is the \textit{search space}, typically box-bounded in $\Re^d$, and $\mathbf{f}:\Omega \rightarrow \Re^M$ is the vector-valued function of the multiple objectives. In MO, due to the conflicting nature of the objectives, it does not exists a unique solution $\mathbf{x^*} \in \Omega$ to the problem (\ref{eq:mo}). The final aim is to identify a set of equally efficient trade-offs among the objectives. This set of efficient trade-offs can be depicted within the space spanned by the $M$ conflicting objectives, allowing for drawing the so-called \textbf{Pareto front} (aka \textit{frontier} or \textit{boundary}). The associated set of solutions – into the search space – is instead known as \textbf{Pareto set}.
An example is shown in Appendix \ref{sec:A.MO}.

Formally, the Pareto set consists of only \textbf{dominant} (aka \textit{not-dominated}) \textbf{solutions}, where a solution $\mathbf{x}$ is said to dominate another solution $\mathbf{x'}$ if their objectives, respectively $\mathbf{f}(\mathbf{x})=\big(f_1(\mathbf{x}),...,f_M(\mathbf{x})\big)$ and $\mathbf{f}(\mathbf{x'})=\big(f_1(\mathbf{x'}),...,f_M(\mathbf{x'})\big)$, satisfy the following two conditions:
\begin{equation}
    \label{eq:cond1}
    f_m(\mathbf{x})\leq f_m(\mathbf{x'}) \; \forall\; m \in \{1,...,M\}
\end{equation}
\begin{equation}
    \label{eq:cond2}
    \exists\; j \in \{1,...,M\} : f_j(\mathbf{x})<f_j(\mathbf{x'})
\end{equation}

Equation (\ref{eq:cond1}) means that $\mathbf{x}$ is not worse than $\mathbf{x'}$ in all the objectives, and equation (\ref{eq:cond2}) means that $\mathbf{x}$ is strictly better than $\mathbf{x'}$ in at least an objective. The Pareto dominance symbol, $\prec$, is used to synthesize (\ref{eq:cond1}-\ref{eq:cond2}):  $\mathbf{f}(\mathbf{x})\prec \mathbf{f}(\mathbf{x'})$.

If the objectives are black-box, their values can only be known point-wise by querying $\mathbf{f}(\mathbf{x})$ at specific locations. Given all the queries performed so far, the set of \textbf{non-dominated solutions} (respectively, \textbf{outcomes}) is the current approximation of the \textbf{Pareto set} (respectively, \textbf{front}), that is the set of the currently \textbf{dominated solutions} (respectively, \textbf{outcomes}).
If the objectives are also expensive to evaluate, in terms of time or resources, then the problem (\ref{eq:mo}) requires to be solved efficiently, meaning that a good Pareto front/set approximation has to be found within a limited number of queries. Thus, sample-efficiency of BO was the driver of its successful extension to the MO setting (i.e., MOBO), mainly along three different strategies:
\begin{itemize}
    \item \textbf{Scalarization} which maps the vector of all objectives into a scalar parametrized function whose optimizer, computed by a single objective method, can span, as the parameters vary, the whole Pareto set \citep{paria2020flexible,zhang2020random}. The key drawback of scalarization is that it does not consider the geometry of the Pareto front approximation.
    \item Maximization of some index related to the quality of the Pareto front approximation. A common choice is the \textbf{dominated hypervolume indicator}, that is the volume of the region dominated by a Pareto front approximation. Based on this, the hypervolume improvement is commonly used in multi-objective optimization. 
    \item \textbf{Information theoretic based}, which aims at reducing the uncertainty/entropy about the Pareto front \citep{belakaria2019max,suzuki2020multi,belakaria2020uncertainty}, recently also considering the multi-fidelity setting \citep{belakaria2020multi}.
\end{itemize}

This paper focuses on the first two strategies. It is important to remark that, while in “vanilla” BO a probabilistic surrogate model is used to approximate the black-box objective function, almost all the MOBO approaches adopt a probabilistic surrogate model for each one of the objectives, assuming independence among them. This is a reasonable assumption, because in MO the objectives should be competing. Exactly as in BO, every new query contributes to better approximate the objective function -- which is vector-valued in MOBO -- through the update of the probabilistic surrogate model. The second key component of BO is the acquisition function, which deals with the well-known exploration-exploitation dilemma. All the “vanilla” BO acquisition functions can be used in the case of scalarization -- because the multi-objective problem is mapped into a single-objective one -- on the contrary, Expected Hypervolume Improvement (EHVI) is an acquisition function specifically designed for vector-vaued MOBO, basically extending the idea underlying the well-known Expected Improvement (EI) to the multi-objective setting.

In this paper we consider HPO optimization of a classification model, with two different objectives to minimize: the misclassification error (MCE) and the \textit{unfairness} metric known as \textbf{Differential Statistical Parity} (DSP), which is detailed in the next section. Both the objectives are computed through stratified 10-fold cross validation, so they are black-box, expensive, multi-extremal, and possibly noisy (depending on the specific ML algorithm to be optimized or the cross-validation procedure).

\subsection{Differential Statistical Parity as unfairness metric}
There is not a unique definition -- and consequently metric -- of fairness \citep{verma2018fairness}. Instead, different alternatives have been proposed depending on application domains and specific use cases. In this paper we consider the DSP -- which has been also recently considered in \citep{schmucker2020multi}. More specifically, we refer to the standard framework where $F_L$ denotes the true labels for the target feature, $F_S$ is the sensitive feature, and $\widehat{F}_L$ denotes the predicted labels. The \textit{Statistical Parity} (SP) requires that positive predictions are unaffected by the value of the sensitive feature(s), independently of the actual label:
\begin{equation*}
    P\big(\widehat{F}_L=1 | F_S=0\big) = P\big(\widehat{F}_L=1 | F_S=1\big)
\end{equation*}
Finally, DSP is a measure of the violation of the above condition, and it is considered as a measure of unfairness.

\subsection{Multiple information source optimization}
Multiple Information Source Optimization (MISO) aims at searching for the global optimum of a black-box, expensive and multi-extremal function, namely the \textit{ground-truth}, given the possibility to also query less expensive \textit{information sources} which are its approximations. The final goal is to find an optimal solution for the \textit{ground-truth} while satisfying some constraint on the query cost accumulated along the search process, basically by effectively and efficiently using the cheap information sources.
MISO has been defined for single-objective problem: differently from multi-objective, here the subscript is used to denote the information source, where $f_1(\mathbf{x})$ is the ground-truth and $f_s(\mathbf{x})$, with $s \in\{2,...,S\}$, are the cheap sources.

The MISO problem can be formulated as:
\begin{alignat}{2}
    \label{eq:miso_obj}
    \mathbf{x^*} = & \quad \underset{\mathbf{x}\in \Omega}{\arg \min} f_1(\mathbf{x})\\
    \label{eq:miso_constr}
    \text{subject to:} & \sum_{(s,\mathbf{x})\in Z^{1:n}} c_s \leq C_{max}
\end{alignat}

where $Z^{1:n}=\bigg\{\Big(s^{(i)},\mathbf{x}^{(i)}\Big)\bigg\}_{i=1:n}$ denotes the ordered set of source-location pairs sequentially queried, $c_s$ is the cost for querying $f_s(\mathbf{x})$, and $C_{max}$ is the maximum query cost that can be accumulated along the sequential optimization process.

BO has been also successfully extended to deal with MISO problems, where each information source, is individually modelled through a probabilistic surrogate model -- usually a Gaussian Process (GP) -- fitted on the queries performed on that source. Then, all the individual models are combined into a single one, which is used to drive the choice of the next promising source-location pair to query, such as in \citep{ghoreishi2019multi,candelieri2021sparsifying}.

\section{Fang-HPO}
The proposed fair and green HPO task is performed by solving the problem (\ref{eq:miso_obj}-\ref{eq:miso_constr}), but with the scalar objective function replaced by a vector-valued one, that is:
\begin{alignat}{2}
    \mathbf{x^*} = & \quad \underset{\mathbf{x}\in \Omega}{\arg \min} \;\mathbf{f_1}(\mathbf{x})\\
    \text{subject to:} & \sum_{(s,\mathbf{x})\in Z^{1:n}} c_s \leq C_{max}
\end{alignat}

where $\mathbf{f_1}$ is the ground-truth, while all the other cheaper information sources are $\mathbf{f_s}$, with $s\in \{2,...,S\}$.

\subsection{Modelling objectives and information sources}
In FanG-HPO, both objectives and information sources are modelled independently via GP regression \citep{williams2006gaussian,gramacy2020surrogates}.
A GP is a probabilistic regression model whose predictive mean, $\mu(\mathbf{x})$, and uncertainty, $\sigma(\mathbf{x})$, are conditioned on previous observations. A brief introduction to GP regression is provided in the Appendix \ref{sec:A.GPR}.





Thus, at a generic iteration, FanG-HPO fits $S \times M$ GP models, leading to the following set of predictive means and uncertainty functions:
$
    \bigg\{\mu_{sm}(\mathbf{x}),\sigma_{sm}(\mathbf{x})\bigg\}_{\substack{s=1:S,\\m=1:M}}
$

Then, a single GP model is fitted, for each objective, by combining the GPs individually modelling that objective on every information source. In FanG-HPO this operation is performed by following the Augmented Gaussian Process (AGP) approach recently proposed in \citep{candelieri2021sparsifying}. More precisely, a set of indices identifying ``\textit{reliable}'' observations from cheaper sources is computed for each source-objective pair, where ``\textit{reliable}'' means they are not too discrepant with respect the ground-truth:
\begin{equation}
\label{eq:index}
\begin{split}
\mathcal{I}_{sm} = \Big\{ & i: \big|\mu_{1m}(\mathbf{x})-\mu_{sm}\big(\mathbf{x}^{(i)}\big)\big|\leq \alpha \sigma_{1m}\big(\mathbf{x}^{(i)}\big), \\
& \mathbf{x}^{(i)} \in \mathbf{X}_s \Big\}, \forall\; s\neq 1, \forall\; m \in \{1,...,M\}
\end{split}
\end{equation}

where $\alpha$ is a technical parameter to tune reliability of observations from cheap information sources (in \citep{candelieri2021sparsifying} the suggested value is $\alpha=1$).

Then, the observations on the ground-truth are ``augmented'' with those identified by $\mathcal{I}_{sm}$, separately for each objective $m\in\{1,...,M\}$:
\begin{equation}
\mathbf{\widehat X}_m \leftarrow \mathbf{X}_1 \cup \Big\{ \mathbf{x}^{(i)} \in \mathbf{X}_s : i \in \mathcal{I}_{sm}, \forall s\neq1\Big\}
\end{equation}

\begin{equation}
\mathbf{\widehat Y}_{m} \leftarrow \mathbf{Y}_{1[m]} \cup \Big\{ y^{(i)} \in \mathbf{Y}_{s[m]} : i \in \mathcal{I}_{sm}, \forall s\neq1\Big\}
\end{equation}
where $\mathbf{X}_s$ are the locations queried on source $s$ and $\mathbf{\widehat{Y}}_{[m]}$ are the values observed for the objective $m$ and associated to the set $\mathbf{\widehat X}_m$ (i.e., the symbol $[m]$ is the operator selecting only the column $m$ of the $n_s \times M$ matrix $\mathbf{Y}_{sm}$, with $n_s$ the number of queries performed on the source $s$).

Finally, FanG-HPO fits $M$ independent AGPs, with predictive means and uncertainty respectively denoted with $\widehat{\mu}_m(\mathbf{x})$ and $\widehat{\sigma}_m(\mathbf{x})$, and both conditioned to $\Big\{\mathbf{\widehat{X}}_m,\mathbf{\widehat{Y}}_m\Big\}$.

\subsection{Deriving the next query}
The next source-location pair to query, namely $(s',\mathbf{x}')$, is derived by solving a multi-objective problem whose objectives are approximated by the AGPs obtained as previously described.
Having $M$ independent AGPs is in line with recent results in literature: as reported in \citep{zhan2017expected} considering a GP modelling each objective independently makes easy the implementation of multi-objective optimization approaches, while using dependent GP models – such as multi output GPs – do not provide any relevant benefit against independent GPs \citep{svenson2016multiobjective}

More precisely, $(s',\mathbf{x}')$ is obtained according to the following two-steps procedure:

\begin{enumerate}
    \item 
    \underline{\textbf{Selecting} $\mathbf{x}'$}. First, the location $\mathbf{x}'$ is selected depending on the well-known Expected Hypervolume Improvement (EHVI). Hypervolume Improvement (HVI) is defined as the relative increase in the hypervolume indicator, when an outcome $\mathbf{y}$, associated to a solution $\mathbf{x}$, is added to the current Pareto front approximation. In BO, the HVI is a random variable because $\mathbf{y}$ is a (set of) random variable itself, and this leads to the EHVI. 
    \begin{equation}
        \mathbf{x}' = \underset{\mathbf{x}\in\Omega}{\arg \max} \; \text{EHVI}(\mathbf{x},\mathcal{P},\mathbf{r})
    \end{equation}
    where $\mathcal{P}$ is the current approximated Pareto front and $\mathbf{r}$ is the \textbf{reference point}. In this paper $\mathbf{r}$ is the worst point, with both MCE and DSP equal to 1.  
    Although a closed formula for the EHVI exists \citep{feliot2017bayesian}, it is expensive to calculate. In FanG-HPO the fast calculation proposed in \citep{zhao2018fast} is used, that is an extension, to the EHVI computation, of the Walking Fish Group (WFG) technique \citep{while2011fast}, one of the fastest algorithms for calculating the hypervolume of a Pareto front approximation.
    
    \item
    \underline{\textbf{Selecting} $s'$}. Then, the information source $s'$ is selected according to both its query cost and its discrepancy with respect to the ground-truth at $\mathbf{x}'$, with respect to all the objectives, that is:
    \begin{equation}
        \label{eq:s'}
        s' = \underset{s\in \{1...,S\}}{\arg \min} \; c_s \cdot \sum_{m=1}^M \Big|\mu_{1m}(\mathbf{x}')-\mu_{sm}(\mathbf{x}')\Big| 
    \end{equation}
\end{enumerate}

Contrary to other recent approaches which propose to query the ground-truth on a regular basis, such as in \citep{khatamsaz2020efficient}, at each iteration FanG-HPO adaptively chooses among all the sources, including the ground-truth.
However, just to ensure a sufficient quality of the approximation provided by the AGPs, before solving (\ref{eq:s'}) FanG-HPO checks if the number of augmenting observations coming from anyone of the cheap sources is larger than those from the ground-truth: in that case $s'=1$ is selected, instead of solving (\ref{eq:s'}).

\section{Experiments and results}

\subsection{Experimental setting}
Experiments consider two benchamark datasets on fairness -- ADULT and COMPAS -- and two ML algorithms whose hyperparameters are optimized -- a Multi-layer Perceptron (MLP) and XGBoost (XGB). Dimensionality of search space, for the two ML algorithms are $d=10$ and $d=7$, respectively, for MLP and XGB. The search spaces are those used in \citep{schmucker2020multi} and reported in Appendix \ref{sec:A.HPO}.

The bi-objective ground-truth is given by the MCE and DSP, computed through stratified 10 fold-cross validation, using the entire datasets. Since DSP is computed for every sensitive feature, we have decided to consider DSP$=\underset{i\in F_S}{\max}\{DSP_i\}$, where $F_S$ is the set of sensitive features and $DSP_i$ is the feature-specific DSP value.

Querying the \textbf{cheap information sources} consists in \textbf{computing the same metrics by using only half of the original dataset}.

According to preliminary evaluations on a set of random configurations of the hyperparameters, the cost for querying the ground-truth is approximately twice that for querying the cheap information sources, for all the datasets and ML algorithms pairs. Therefore, we set $c_1=2$ and $c_2=1$.

It is important to remark that only FanG-HPO exploits the two information sources, while the HPO task performed through BoTorch uses only the ground-truth. We have selected BoTorch as a baseline because it provides implementations of many state-of-the-art algorithms for \textit{vanilla} BO, multi-objective BO, and multi-fidelity BO. Testing all of them is out of the scope of this paper. Thus, we have selected one of the most effective and efficient implementations for multi-objective (single-source) BO, that is ParEGO \citep{knowles2006parego} with $q$-Knowledge Gradient ($q$KG) as acquisition function \citep{wu2016parallel}. Specifically, we have used $q=5$ in our experiments. ParEGO is an extension of the (single-objective) Efficient Global Optimization (EGO) algorithm \citep{jones1998efficient} to the MO setting, whose core consists in approximating the different objectives through independent GPs and use scalarization to rescale the multi-objective problem in to a single-objective one. It is also important to remark that -- as all the other available BO tools -- BoTorch considers multi-objective and multi-fidelity optimization as two separate problems, so it does not provide any algorithm to jointly address them. Moreover, BoTorch provides implementations for multi-fidelity optimization, which is just a special case of multiple information source optimization, and consequently we have discarded them comparison because not well-suited for a comparison. 
\color{black}

First, $2d$ hyperparameters are randomly chosen to generate the initial designs for MLP and XGB, separately. For BoTorch this leads to an initialization cost of $2dc_s$, that is $40$ and $28$ for MLP and XGB (independently on the dataset, in this study). The maximum accumulated query cost has been set to $C_{max}=20d$, that is $C_{max}=200$ and $C_{max}=140$ for MLP and XGB, respectively.

To have a fair comparison between BoTorch and FanG-HPO, $13$ and $9$ hyperparameters configurations are sampled from the BoTorch initial designs, respectively for MLP and XGB, leading to an associated query cost of $26$ and $18$. The remaining budgets for initialization (i.e., $40-26=14$ and $28-18=10$) are used to generate and evaluate initial designs for the cheap information sources, that are $14$ hyperparameters configurations for MLP and $10$ for XGB.

Finally, to mitigate the randomness due to (\textit{i}) generation of initial design and (\textit{ii}) MLP and XGBoost learning algorithms, we have generated five initial designs -- for each ML algorithm -- and performed five independent runs for each one of them.

The hypervolume (HV), with respect to the accumulated query cost, is the performance metric used to monitor the effectiveness of the two BO-based approaches. As already mentioned, the reference point is the worst one, that is MCE=1 and DSP=1.

Moreover, we have also considered two fairness-aware algorithms, specifically \textbf{zlrm} and \textbf{fgrrm} \citep{scutari2021achieving}. To mitigate the effect of randomness, five independent runs have been performed for both the FairML algorithms. As usual, a constraint on the unfairness must be provided for these algorithms: we set this value to 0.1 (for each individual sensitive feature).

FanG-HPO was developed in R and integrates, through the \textbf{reticulate} R package, Python code (i.e., sklearn modules for MLP and XGBoost). FairML algorithms are from the \textbf{fairml} R package.

\subsection{Experimental results}

\subsubsection{FanG-HPO vs BoTorch-based HPO}
\begin{figure*}[ht]
    \vskip 0.2in
    \begin{center}
        \centerline{
            \includegraphics[width=0.25\textwidth]{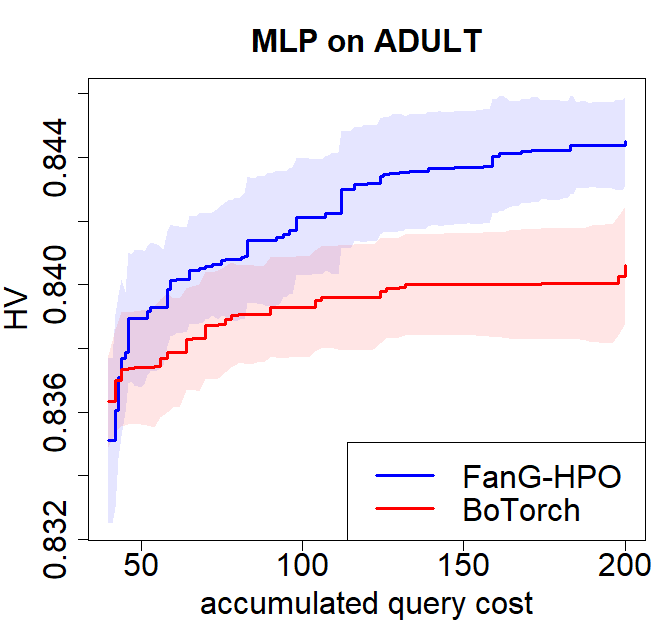}
            \includegraphics[width=0.25\textwidth]{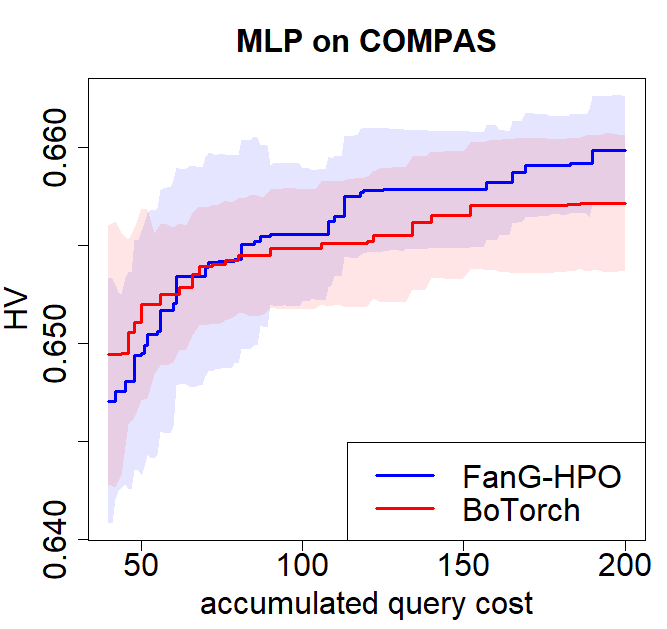}
            \includegraphics[width=0.25\textwidth]{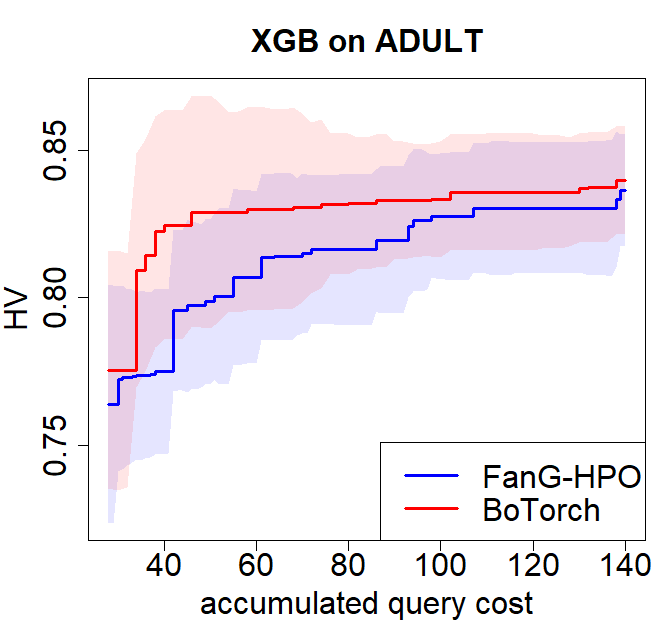}
            \includegraphics[width=0.25\textwidth]{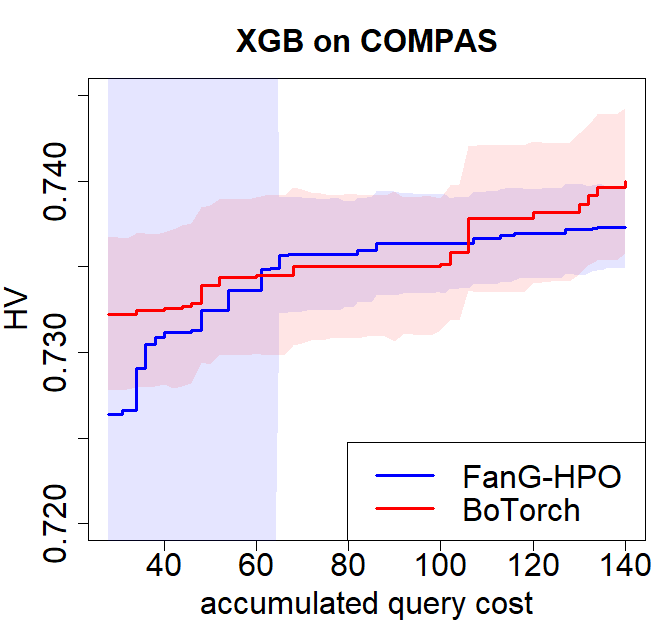}
        }
        \caption{Hypervolume (HV) with respect to query cost accumulated over the HPO process (median $\pm$ standard deviation over the 25 experiments: 5 initial designs times 5 repeated runs for each design). Comparison between FanG-HPO and BoTorch based HPO.}
        \label{fig:hv}
    \end{center}
    \vskip -0.2in
\end{figure*}

This section summarizes the most relevant results of the study. Figure \ref{fig:hv} compares FanG-HPO and BoTorch-based HPO according to the evolution of the HV, associated to the current Pareto front approximation, with respect to the query cost accumulated over the HPO process. First, it is important to remark two important points: (\textit{i}) only observations on the ground-truth are used to compute the approximate Pareto front at each iteration (and consequently the associated HV), also for FanG-HPO, and (\textit{ii}) the small initial difference between the HV of the two methods is due to the fact that FanG-HPO uses just a subsample of the initial design (on the ground-truth) of BoTorch (i.e., FanG-HPO must split the initial budget to initialize the GPs on both the ground-truth and the cheap information source).

As far as HPO of MLP is concerned, FanG-HPO outperforms BoTorch-based HPO. More specifically, given the same cost -- but a different number of hyperparameters configurations evaluated -- FanG-HPO identifies MLP models offering a better trade-off between MCE and DSP. Indeed, the HV -- computed as the median on the 25 runs -- is higher for FanG-HPO almost immediately in the case of the ADULT dataset and after half of the available budget in the case of COMPAS (i.e., accumulated query cost equal to $C_{max}/2=100$). This means that, given a desired accumulated query cost -- which is a proxy for computational burden and, consequently, energy consumption -- FanG-HPO identifies ML models which are more Pareto-efficient (i.e., higher HV) than those generated via BoTorch. Analogously, given a desired level of Pareto-efficiency (i.e., a HV value), FanG-HPO achieves it with a lower computational burden (i.e., accumulated query cost). Therefore, FanG-HPO is more effective and \textit{greener} than BoTorch-based HPO.

As reported in Table \ref{table:results}, at the end of the optimization process the difference in terms of HV is statistically significant, with a confidence level $0.05$ (Mann-Whitney's U test, null hypothesis: final HVs are similar between the two approaches; alternative hypothesis: final HV for FanG-HPO is higher than for BoTorch).
\begin{table*}[t]
    \caption{Hypervolumes (median$\pm$sd over 25 independent runs) of the final Pareto front approximations provided by FanG-HPO and BoTorch.}
    \label{table:results}
    \vskip 0.15in
    \begin{center}
        \begin{small}
            \begin{sc}
                \begin{tabular}{lcccr}
                \toprule
                HPO task & Fang-HPO & BoTorch & $p$-value \\
                \midrule
                MLP on ADULT & \textbf{0.8445} $\pm$ \textbf{0.0014} & 0.8406 $\pm$ 0.0018 & \textbf{$<$0.0001} \\
                MLP on COMPAS & \textbf{0.6598} $\pm$ \textbf{0.0028} & 0.6571 $\pm$ 0.0035 & \textbf{0.0479} \\
                XGBoost on ADULT & 0.8363 $\pm$ 0.0190 & \textbf{0.8397} $\pm$ \textbf{0.0183} & 0.4163 \\
                XGBoost on COMPAS  & 0.7373 $\pm$ 0.0024 & \textbf{0.7400} $\pm$ \textbf{0.0042} & 0.9463 \\
                \bottomrule
                \end{tabular}
            \end{sc}
        \end{small}
    \end{center}
    \vskip -0.1in
\end{table*}

Results are not equally encouraging when HPO on XGBoost is considered. A possible motivation is that XGBoost is an effective ML algorithm able to provide accurate and fair models by itself, so there are not significant differences in optimizing its hyperparameters through FanG-HPO or BoTorch-based HPO. This is also confirmed by the results in Table \ref{table:results}, where the slightly difference between the two approaches, in terms of HV at the end of the optimizaiton, is not statistically significant (Mann-Whitney's U test, null hypothesis: final HVs are similar between the two approaches; alternative hypothesis: final HV for BoTorch is higher than for FanG-HPO). In the case of the ADULT dataset, the HV of BoTorch-based HPO is, on median, higher than the FanG-HPO's one. However, the difference quickly decreases with the accumulated query cost and becomes not statistically significant (Figure \ref{fig:hv}). In the case of the COMPAS dataset, the two approaches are comparable, with FanG-HPO slightly better, on median, when the accumulated query cost is, approximately, within the range $[60;110]$, and then BoTorch slightly better for higher accumulated query costs. The initially large standard deviation for FanG-HPO (around 0.05) is due to the random subsampling of 9 (out of the 14) hyperparameters configurations from the initial design of BoTorch, performed as explained in the experimental setting. Usage of the cheap source depends on the run -- basically on the initialization of the design. 

\begin{figure*}[h!]
    \vskip 0.2in
    \begin{center}
        \centerline{
            \includegraphics[width=0.4\textwidth]{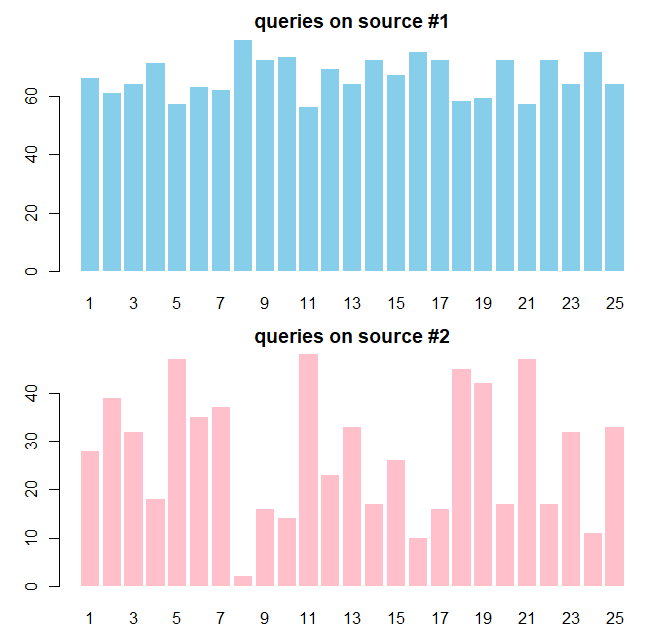}
            \hspace{1.3cm}
            \includegraphics[width=0.4\textwidth]{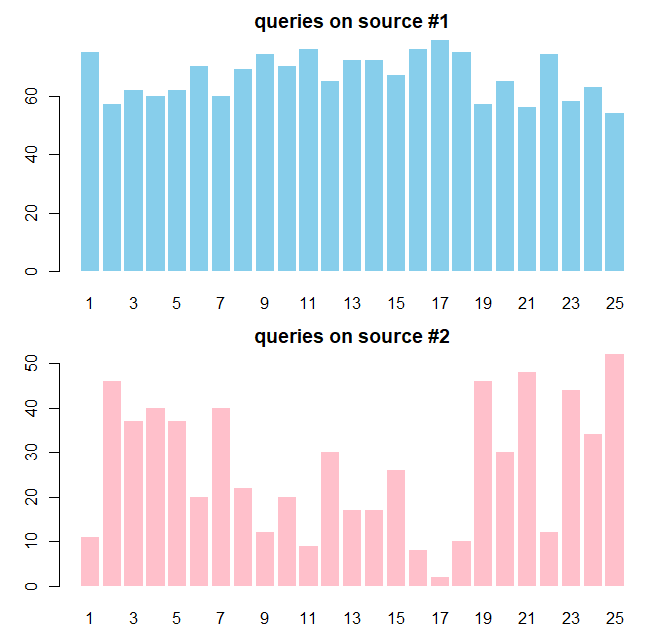}
        }
        \caption{Number of queries performed by FanG-HPO on each source for optimizing: (left) MLP on the ADULT dataset, and (right) MLP on the COMPAS dataset.}
        \label{fig:queries1}
    \end{center}
    \vskip -0.2in
\end{figure*}

\begin{figure*}[h!]
    \vskip 0.2in
    \begin{center}
        \centerline{
            \includegraphics[width=0.4\textwidth]{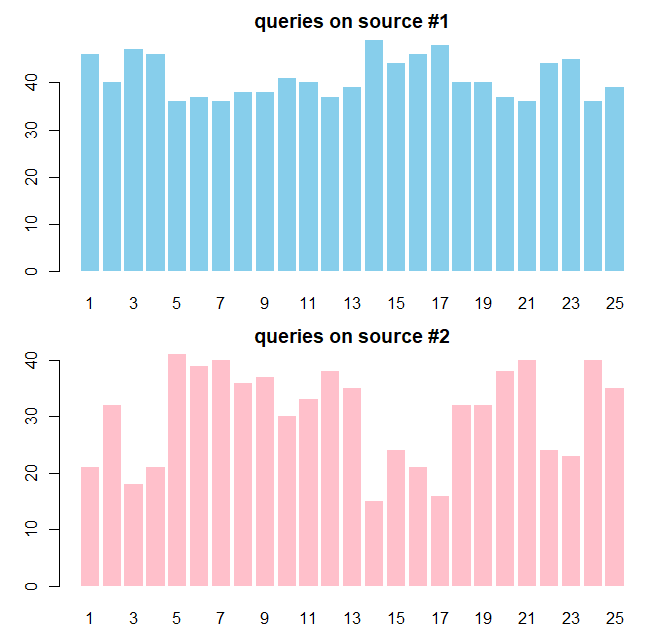}
            \hspace{1.3cm}
            \includegraphics[width=0.4\textwidth]{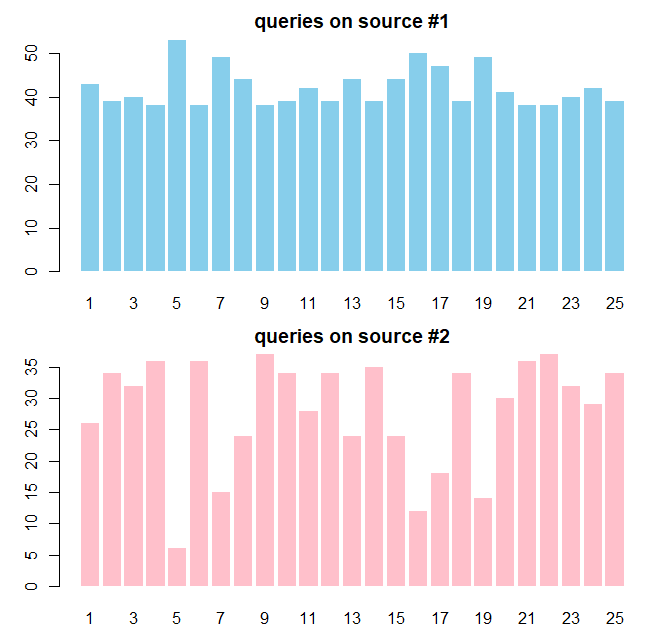}
        }
        \caption{Number of queries performed by FanG-HPO on each source for optimizing: (left) XGBoost on the ADULT dataset, and (right) XGBoost on the COMPAS dataset.}
        \label{fig:queries2}
    \end{center}
    \vskip -0.2in
\end{figure*}

Summarizing, FanG-HPO does not apparently underperfom BoTorch-based HPO and, in some cases, can significantly outperform it. This achievement is obtained by how FanG-HPO exploits information from the cheap information source. In Figure \ref{fig:queries1} and Figure \ref{fig:queries2}, we have reported the number of queries of FanG-HPO, on each information source and for each one of the 25 runs, separately for each ML algorithm and dataset pair. The queries related to the initial designs are not included in the counts.

Although the encouraging results, only a larger set of experiments -- performed also by other research groups -- involving more datasets and ML algorithms could definitely confirm these preliminary results. 

\subsubsection{Comparison against fairML algorithms}
Finally, we present a Pareto analysis of the solutions provided by both FanG-HPO and BoTorch-based HPO against the two fairml algorithms \textbf{zlrm} and \textbf{fgrrm}.
Separately for the two datasets and the two ML algorithms (i.e., MLP and XGBoost), we have retrieved the best Pareto fronts for FanG-HPO and BoTorch-based HPO (i.e., the ones with the largest HV over all the 25 independent runs). A separate chart has been obtained for each dataset, depicting also the MCE and DSP values provided by the five indepent runs of \textbf{zlrm} and \textbf{fgrrm}. Pareto fronts are zoomed in for a better visualization, so the reference point (i.e., MCE=1 and DSP=1) is not in the charts.

\begin{figure}[h]
    \begin{center}
        \centerline{
            \includegraphics[width=0.47\textwidth]{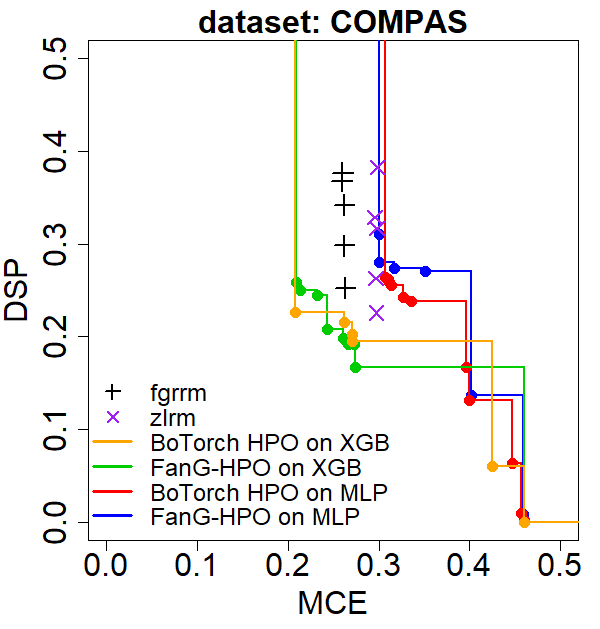}
        }
        \caption{Pareto analysis of the models identified by the different approaches on the COMPAS dataset.}
        \label{fig:pareto_compas}
    \end{center}
    \vskip -0.2in
\end{figure}

\begin{figure}[h]
    \begin{center}
        \centerline{
            \includegraphics[width=0.47\textwidth]{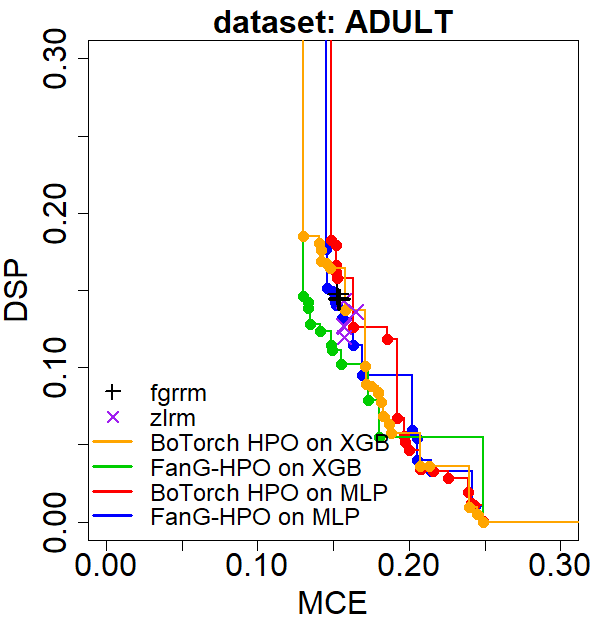}
        }
        \caption{Pareto analysis of the models identified by the different approaches on the ADULT dataset.}
        \label{fig:pareto_adult}
    \end{center}
    \vskip -0.2in
\end{figure}

Figure \ref{fig:pareto_compas} shows the results for the COMPAS dataset. Important remarks are:
\begin{itemize}
    \item XGBoost models dominate -- in Pareto terms -- most of the MLP models, irrespectively to the adoption of FanG-HPO or BoTorch. However, some not-dominated MLP models offer a lower DSP (in the face of an increase in MCE);
    \item fairml models identified through zlrm and fgrrm dominate many MLP models but are completely dominated by XGBoost models;
    \item the Pareto fronts identified by FanG-HPO and BoTorch are quite similar, as well as the number of Pareto efficient models. It is important to remark that solutions on the Pareto front refer to the ground-truth only, but FanG-HPO performs a lower number of queries on it.
\end{itemize}
Figure \ref{fig:pareto_adult} shows results on ADULT. Important remarks are:
\begin{itemize}
    \item XGB models identified by FanG-HPO dominate most of the other XGB and MLP models, as well as all the models identified through zlrm and fgrrm;
    \item MLP models identified by FanG-HPO dominate most of the MLP models identified by BoTorch-based HPO;
    \item zlrm and fgrrm models lay on the approximated Pareto front of FanG-HPO on MLP, and dominate a portion of that of BoTorch-based HPO.
\end{itemize}

\section{Conclusions and perspectives}
The proposed approach, namely \textbf{FanG-HPO} -- \textbf{\underline{F}}air \textbf{\underline{an}}d \textbf{\underline{G}}reen hyperparameter optimization -- is aimed to searching for accurate and fair ML models while using a small portions of the large dataset to also reduce computational time for training and validation and, consequently energy consumption and possibly associated CO$_2$ emissions.

Preliminary results proved that FanG-HPO does not underperform multi-objective (i.e., accuracy and fairness) HPO built on BoTorch. Furthermore, the capability to deal with multiple information sources (i.e., small portions of the largest dataset) allows FanG-HPO to significantly improve efficiency in terms of accumulated query costs, compared to BoTorch. This means that an approximate Pareto front with a largest hypervolume can be identified with a lower accumulated computational time required to train-and-validate ML models. This was clearly observed in the case of HPO of MLP, on both ADULT and COMPAS datasets.

Finally, depending on the ML algorithm to be optimized, both FanG-HPO and BoTorch-based HPO can identify more efficient models -- in Pareto terms -- than fairness-aware ML algorithms, specifically zlrm and fgrrm.

Ongoing and future works will focus on a larger set of experiments, including more ML algorithms and fairness datasets, and on investigating the possibility to also consider \textit{cost-aware optimization}, recently proposed in \citep{lee2020cost,candelieri2021miso,luong2021adaptive}, where sources' query costs are not fixed but depends on the hyperparameters configuration to evaluate. Although the two-steps acquisition function proposed in FanG-HPO should not require any awareness about location-dependent costs (i.e., after choosing $x'$ the query costs only depends on the size of dataset underlying the information sources), it could be anyway interesting to investigate this topic.


\section*{Software and Data}
Both the R code and the data are available on request.


\section*{Acknowledgements}
We greatly acknowledge the DEMS Data Science Lab, Department of Economics Management and Statistics (DEMS), University of Milano-Bicocca, for supporting this work by providing computational resources.


\nocite{langley00}

\bibliography{FanG-HPO.bib}
\bibliographystyle{icml2022}

\newpage
\appendix
\onecolumn
\section{Appendix}
\label{sec:appendix}

\subsection{Multi-objective optimization}
\label{sec:A.MO}
\begin{figure*}[ht]
    \vskip 0.2in
    \begin{center}
        \centerline{\includegraphics[width=0.9\textwidth]{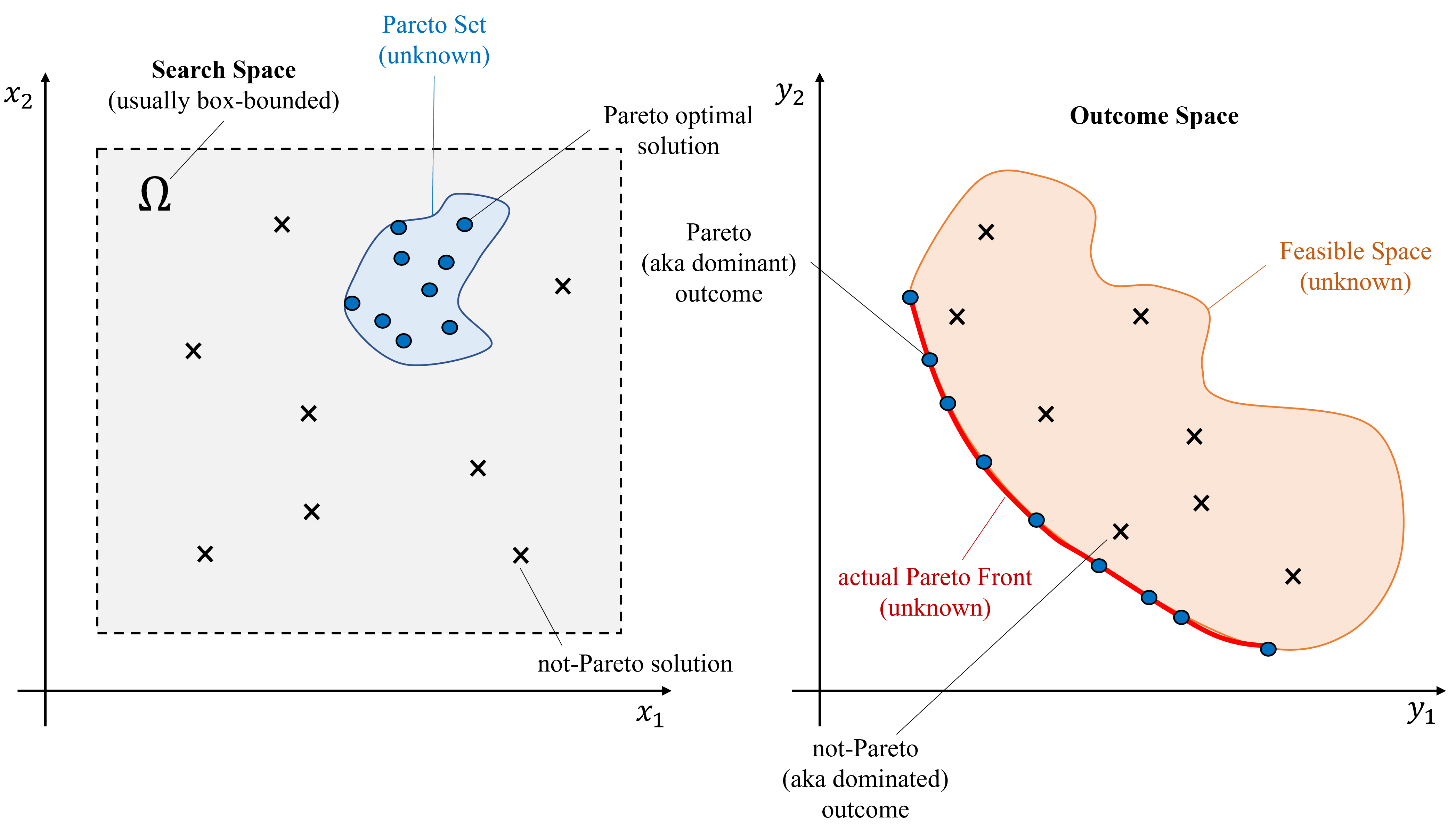}}
        \caption{On the left: box-bounded Search Space $\Omega$, (unknown) Pareto Set, Pareto and not-Pareto solutions. On the right: (unknown) Feasible Space within the Outcome Space (i.e., consisting of all the outcomes associated to solutions in the Search Space), Pareto (aka dominant) outcomes, not-Pareto (aka dominated outcomes), and actual (unknown) Pareto Front.}
        \label{fig:pareto}
    \end{center}
    \vskip -0.2in
\end{figure*}

Figure \ref{fig:pareto} summarizes the main concepts of Pareto analysis in the multi-objective setting. For the sake of visualization, a two objectives problem, with a two dimensional search space $\Omega$, is considered. On the left hand side the search space and the (unknown) Pareto set are depicted; on the right hand side, the outcome space spanned by the two objectives is illustrated, along with the (unknown) feasible space (containing the outcomes associated to all the possible solutions in the search space) and the (unknown) actual Pareto front.

\subsection{Gaussian Process Regression}
\label{sec:A.GPR}
A Gaussian Process (GP) is a collection of random variables, any finite number of which have a joint Gaussian distribution, and it is completely specified by its mean function, $\mu(\mathbf{x})$, and covariance function, $k(\mathbf{x},\mathbf{x'})$. A GP is denoted with $\mathcal{GP}\big(\mu(\mathbf{x}),k(\mathbf{x},\mathbf{x'})\big)$ \citep{williams2006gaussian,gramacy2020surrogates}. Importantly, these two scalar-valued functions can be conditioned on a set of available observations, leading to a \textit{probabilistic regression model} which can be used to make predictions at any location $\mathbf{x}$, according to the so-called GP's \textit{predictive} (aka \textit{posterior}) mean and standard deviation. While the first represents the predicted value, the second represents the associated predictive uncertainty.

Consider to have performed $n$ queries, then denote with $\mathbf{X}^{1:n}=\big\{\mathbf{x}^{(i)}\big\}_{i=1:n}$ the set of queried locations and with $\mathbf{Y}^{1:n}=\big\{y^{(i)}\big\}_{i=1:n}$ the associated observed outcomes, possibly noisy (i.e., $y^{(i)}=f\big(\mathbf{x}^{(i)}\big)+\varepsilon^{(i)}$, where $\varepsilon^{(i)}$ is assumed to be a zero-mean Gaussian noise, $\varepsilon^{(i)}\sim\mathcal{N}\big(0,\sigma^2_\varepsilon\big), \forall\; i\in\{1,...,n\}$).
Then, the GP's predictive mean and variance, conditioned to the $n$ performed queries, are respectively computed as follows:

\begin{equation}\label{eq:mu}
	\mu(\mathbf{x})=\mathbf{k}\big(\mathbf{x},\mathbf{X}^{1:n}\big)\big[\mathbf{K} - \sigma^2_\varepsilon\mathbf{I}\big]^{-1}\mathbf{y}
\end{equation}

\begin{equation}\label{eq:sd}
	\sigma^2(\mathbf{x})=k\big(\mathbf{x},\mathbf{x}\big) -\mathbf{k}\big(\mathbf{x},\mathbf{X}^{1:n}\big)\big[\mathbf{K} - \sigma^2_\varepsilon\mathbf{I}\big]^{-1} \mathbf{k}\big(\mathbf{X}^{1:n},\mathbf{x}\big)
\end{equation}

where $\mathbf{K}$ is an $n \times n$ matrix whose entries are $k_{ij}=k\big(\mathbf{x}^{(i)},\mathbf{x}^{(j)}\big)$, and $\mathbf{k}(\mathbf{x},\mathbf{X}^{1:n})$ is the $n$-dimensional row vector $\big( k(\mathbf{x},\mathbf{x}^{(1)}), ..., k(\mathbf{x},\mathbf{x}^{(n)})\big)$. Just for completeness, $\mathbf{k}\big(\mathbf{X}^{1:n},\mathbf{x}\big)$ is the $n$-dimensional column vector $\mathbf{k}\big(\mathbf{x},\mathbf{X}^{1:n}\big)^\top$.
The GP's predictive uncertainty is the posterior standard deviation, that is $\sigma(\mathbf{x})=\sqrt{\sigma(\mathbf{x})^2}$.

Before conditioning a GP to a set of observations, two priors must be provided, relatively to the mean and the covariance functions. Usually, the first is set to zero: this is not a limitation because the posterior mean will be not confined to this value. However, it is possible to incorporate explicit basis functions for expressing prior information on the function to approximated by the GP model~\citep{williams2006gaussian}.
On the contrary, the covariance function is chosen among a set of possible \textit{kernel functions}, such as Squared Exponential, Power Exponential, and Mat\'ern kernels, offering different modelling options with respect to structural properties of the function to be approximated, especially \textit{smoothness}~\citep{gramacy2020surrogates,archetti2019bayesian,frazier2018bayesian,williams2006gaussian}.

\subsection{HPO search spaces}
\label{sec:A.HPO}

\begin{table*}[h]
    \caption{sklearn MLP's search space.}
    \label{table:results}
    \vskip 0.15in
    \begin{center}
        \begin{small}
            \begin{sc}
                \begin{tabular}{lccc}
                \toprule
                Hyerparameter & Type & Domain & Scaling \\
                \midrule
                n\_layers & integer & \{1,2,3,4\} & linear \\
                layer\_1 & integer & \{2,...,32\} & log$_2$ \\
                layer\_2 & integer & \{2,...,32\} & log$_2$ \\
                layer\_3 & integer & \{2,...,32\} & log$_2$ \\
                layer\_4 & integer & \{2,...,32\} & log$_2$ \\
                alpha & real & [10$^{-6}$,10$^{-1}$] & log$_{10}$ \\
                learning\_rate\_init & real & [10$^{-6}$,10$^{-1}$] & log$_{10}$ \\
                beta\_1 & real & [0.001,0.99] & log$_{10}$ \\
                beta\_2 & real & [0.001,0.99] & log$_{10}$ \\
                tol & real & [10$^{-5}$,10$^{-2}$] & log$_{10}$ \\
                \bottomrule
                \end{tabular}
            \end{sc}
        \end{small}
    \end{center}
    \vskip -0.1in
\end{table*}

\begin{table*}[h]
    \caption{sklearn XGBoost's search space.}
    \label{table:results}
    \vskip 0.15in
    \begin{center}
        \begin{small}
            \begin{sc}
                \begin{tabular}{lccc}
                \toprule
                Hyerparameter & Type & Domain & Scaling \\
                \midrule
                n\_estimators & integer & \{1,...,256\} & log$_2$ \\
                learning\_rate & real & [0.01,1.0] & log$_{10}$ \\
                gamma & real & [0.0,0.1] & linear \\
                reg\_alpha & real & [10$^{-3}$,10$^3$] & log$_{10}$ \\
                reg\_alpha & real & [10$^{-3}$,10$^3$] & log$_{10}$ \\
                subsample & real & [0.01,1.0] & linear \\
                max\_depth & integer & \{1,2,...,16\} & linear \\
                \bottomrule
                \end{tabular}
            \end{sc}
        \end{small}
    \end{center}
    \vskip -0.1in
\end{table*}


\end{document}